\documentclass[letterpaper]{article} % DO NOT CHANGE THIS
\usepackage{aaai23}  % DO NOT CHANGE THIS
\usepackage{times}  % DO NOT CHANGE THIS
\usepackage{helvet}  % DO NOT CHANGE THIS
\usepackage{courier}  % DO NOT CHANGE THIS
\usepackage[hyphens]{url}  % DO NOT CHANGE THIS
\usepackage{graphicx} % DO NOT CHANGE THIS
\usepackage{natbib}  % DO NOT CHANGE THIS AND DO NOT ADD ANY OPTIONS TO IT
\usepackage{caption} % DO NOT CHANGE THIS AND DO NOT ADD ANY OPTIONS TO IT
\usepackage{algorithm}
\usepackage{algorithmic}

\usepackage{newfloat}
\usepackage{listings}
\DeclareCaptionStyle{ruled}{labelfont=normalfont,labelsep=colon,strut=off} % DO NOT CHANGE THIS
\floatstyle{ruled}
\newfloat{listing}{tb}{lst}{}
\floatname{listing}{Listing}
\title{Coupling Artificial Neurons in BERT and Biological Neurons in the Human Brain}
\author {
    Xu Liu\textsuperscript{\rm 1}\equalcontrib,
    Mengyue Zhou\textsuperscript{\rm 1}\equalcontrib, 
    Gaosheng Shi\textsuperscript{\rm 1}\equalcontrib, 
    Yu Du\textsuperscript{\rm 1}, 
    Lin Zhao\textsuperscript{\rm 2}, 
    Zihao Wu\textsuperscript{\rm 2}, 
    David Liu\textsuperscript{\rm 3}, 
    Tianming Liu\textsuperscript{\rm 2},
    Xintao Hu\textsuperscript{\rm 1}\thanks{The corresponding author.}
}
\affiliations {
    \textsuperscript{\rm 1} School of Automation, Northwestern Polytechnical University\\
    \textsuperscript{\rm 2} School of Computing, University of Georgia\\
    \textsuperscript{\rm 3} Athens Academy \\
    \{liu\_xu, zhou\_my, 2021202420, dddyyy\}@mail.nwpu.edu.cn, \{lin.zhao, zw63397\}@uga.edu, \{david.weizhong.liu, tianming.liu\}@gmail.com, xhu@nwpu.edu.cn
}

\usepackage{bibentry}
\begin{document}

\maketitle

\begin{abstract}
    Linking computational natural language processing (NLP) models and neural responses to language in the human brain on the one hand facilitates the effort towards disentangling the neural representations underpinning language perception, on the other hand provides neurolinguistics evidence to evaluate and improve NLP models. Mappings of an NLP model’s representations of and the brain activities evoked by linguistic input are typically deployed to reveal this symbiosis. However, two critical problems limit its advancement: 1) The model’s representations (artificial neurons, ANs) rely on layer-level embeddings and thus lack fine-granularity; 2) The brain activities (biological neurons, BNs) are limited to neural recordings of isolated cortical unit (i.e., voxel/region) and thus lack integrations and interactions among brain functions. To address those problems, in this study, we 1) define ANs with fine-granularity in transformer-based NLP models (BERT in this study) and measure their temporal activations to input text sequences; 2) define BNs as functional brain networks (FBNs) extracted from functional magnetic resonance imaging (fMRI) data to capture functional interactions in the brain; 3) couple ANs and BNs by maximizing the synchronization of their temporal activations. Our experimental results demonstrate 1) The activations of ANs and BNs are significantly synchronized; 2) the ANs carry meaningful linguistic/semantic information and anchor to their BN signatures; 3) the anchored BNs are interpretable in a neurolinguistic context. Overall, our study introduces a novel, general, and effective framework to link transformer-based NLP models and neural activities in response to language and may provide novel insights for future studies such as brain-inspired evaluation and development of NLP models.
\end{abstract}

\section{Introduction}

Linking computational natural language processing (NLP) models and neural responses to language in the human brain has attracted increasing interest recently. On the one hand, computational NLP models facilitate the effort towards disentangling the neural representations and cortical structures underpinning language perception \cite{caucheteux2021language,schrimpf2020integrative,goldstein2020thinking}. On the other hand, researchers have also attempted to leverage data and evidence from neurolinguistics to evaluate and improve NLP models \cite{schwartz2019inducing,toneva2019interpreting,marvin2018targeted}. 

The relationship between NLP models and neural responses is typically established by building a mapping between the feature space and the brain activity space. The feature space is spanned by the featural representations in NLP models, referred to as the responses of artificial neurons (ANs). The brain activity space describes the quantitative brain activities extracted from functional neuroimaging data, referred to as activations of biological neurons (BNs). 

The successful deployment of such a framework leads to fruitful discoveries in both neurolinguistics and computational NLP modeling. However, two critical problems remain: 1) The feature space is lacking in fine-granularity. Most existing studies adopt layer-level embeddings as NLP features. As argued in previous studies, a fine decomposition of a model’s components and measurement of their internal representations and operations are among the keys for mapping the elementary units of NLP to their neurobiological counterparts\cite{doi:10.1146/annurev-linguistics-051421-020803}; 2) The brain activity space lacks of information about the integration and interaction of brain functions. Most existing studies treat each voxel or brain region as an independent brain activity unit, ignoring the complex regional/systematical interactions (coactivations or activation-deactivation) in the human brain that have been widely observed in the process of language perception \cite{xiong2021both,Saurabh2019Naturalistic}.

We sought to address the following questions: 
\begin{itemize}
	\item How to define the ANs with fine-granularity in transformed-based NLP models? How can their activations be quantified?
	\item Do those ANs carry meaningful linguistic/semantic information?
	\item Do those ANs anchor to their BN signatures that can reveal functional interaction in the brain? Are the BNs they anchored interpretable in a neurolinguistic perspective?
\end{itemize}

To this end, we propose a general framework for coupling the ANs in transformer-based NLP models and the BNs in the human brain.  We adopt the “Narrative” fMRI dataset \cite{nastase2021narratives} which were acquired while human subjects listening to naturalistic spoken stories to implement the framework. In brief, we use the pre-trained BERT model \cite{devlin2018bert} to embed the transcript of the stories. We define each hidden dimension in the multi-head self-attention module as a single AN. The temporal activation of an AN is quantified according to the element-wise product of the queries and keys. We then adopt the fMRI data decomposition model based on sparse deep belief network to identify BNs in the human brain and retrieve their temporal activations. The coupling between ANs and BNs is achieved by maximizing the correlations between their temporal activations. Finally, we provide neurolinguistic interpretation of the coupled AN-BN pairs in terms of part-of-speech tags. 

Our experimental results show: 1) The ANs of BERT defined in this study carry meaningful linguistic information; 2) The ANs well synchronize to BNs and thus anchor to their BN signatures; 3) The coupled AN-BN pairs are interpretable in a neurolinguistic context.  

\section{Related Works}
\subsection{Linking NLP Models and Neural Responses}
Despite the fundamental difference from the neural architectures of the human brain, computational NLP models have been found to show considerable representational alignment to neural responses \cite{caucheteux2021language,schrimpf2020integrative,goldstein2020thinking,fedorenko2020lack}, suggesting that NLP models may serve as potential tools to explore the representation and neural circuits underpinning linguistic cognition. A linear transformation is typically trained to map between computational representations of NLP models and neural responses to the same set of stimuli. Fitness of the model, also known as “brain score”\cite{mitchell2008predicting}, is used to establish the correspondence between them. Here we briefly summarize related works and a comprehensive review is referred to \cite{abdou2022connecting}.

Early studies focused on linking models’ representations (e.g, frequency of co-occurrence and concept-relation-feature triple) of and neural response to isolated word or phrase \cite{mitchell2008predicting,devereux2010using,gauthier2018does,beinborn2019robust}. Later on, researchers used NLP models (e.g., context-free grammars, \cite{levy2008expectation,reitter2011computational}, \textit{n}-gram Markov chain \cite{parviz2011using}, syntactic surprisal estimation \cite{frank2015erp,brennan2019hierarchical}, recurrent neural network grammar (RNNG,\cite{hale2018finding} and subgraph embeddings \cite{reddy2021syntactic}, just name a few) to build syntactic features to explore how the brain represents syntactic structure. NLP models such as auto-regressive RNN \cite{wehbe2014simultaneously}, word embedding built from co-occurrence statistics \cite{huth2016natural} have been used to depict multi-levels of perceptual and linguistic abstraction. 

The recently advanced NLP models have led to both better performance on various linguistics tasks and improved prediction of neural responses. Schrimpf et al. \cite{schrimpf2020integrative} evaluated a wide variety of NLP models, ranging from simple word embedding to the ones built on self-attention, based on their predictiveness of neural response and self-paced reading patterns. Similarly, Caucheteux et al. \cite{caucheteux2021language} tested even a larger model set including 7400 models. Subsequently, the authors in \cite{antonello2021low} described the relationships between representations derived from 100 different NLP models by using an encoder-decoder framework to measure the transferability between different models.  

Researchers have also applied computational controls on models for neurolinguistic studies. Those controls include varying the input context length \cite{jain2018incorporating,abnar2019blackbox}, finetuning a baseline model on a suit of linguistic tasks and comparing the representations before and after finetuning \cite{gauthier2019linking,abdou2021does}, disentangling composed-from individual-word meaning \cite{toneva2022combining} and factorizing distributed representations into specific linguistic factors (e.g., syntax vs. semantics and lexical vs. compositional)\cite{caucheteux2021disentangling}. 

Despite the fruitful outcomes, the existing studies that link self-attention based NLP models to neural responses face two limitations. First, in most studies the model's representation rely on layer-level embeddings. In spite of lacking an explicit definition of ANs in computational models, those studies implicitly treat each layer as a single AN. However, the layer-level representations are derived through complex internal operations and transformations in multi-head self-attention (MSA) module. And "multi-head" itself originates from the multiple types of attentional relationships among input sequence. Thus, regarding a layer as a single AN is lacking in fine-granularity. Defining ANs with fine-granularity and measuring their internal operations are desired to advance the mapping between the elementary units of NLP to their neurobiological counterparts, which is a key objective of this work. Second, the quantification of neural responses (biological neurons, BNs) in exiting studies is limited to isolated units of brain voxles or regions. However, the brain is intrinsically organized in complex networked systems and brain functions essentially depend on functional interactions among voxels/regions. Neural response quantification that can capture inter-regional interactions is expected to disentangle the neurobiological counterparts of NLP models' elementary units.    

\subsection{Interpretation of Transformer-Based NLP Models}
Our study is related to and inspired by the studies that interpret transformer-based NLP models based on visualizations. Different aspects of the model have been visualized, for example, the attention maps\cite{vig2019visualizing,clark2019does,aken2020visbert}, the relationship between attention and model outputs \cite{jain2019attention}, attention flow \cite{abnar2020quantifying}, evolution of representations\cite{voita2019bottom}, the analysis of captured linguistic information via probing \cite{tenney2019BERT}, and multilinguality\cite{dufter2020identifying}. In particular, some visualization tools \cite{vig2019visualizing,clark2019does,aken2020visbert} explore views at different levels of granularity, the attention-head view visualizing the attention patterns generated by one or more attention heads in a given layer, the model view visualizing attention across all of the layers and heads, and the neuron view visualizing the individual dimension in the query and key vectors. The neuron view inspired us to define fine-granularity ANs in transformer-based NLP models, however, measuring the temporal activations of ANs has rarely been formulated, which motivated us to explore new methodologies in this work.

\subsection{Functional Brain Networks in fMRI}
Functional brain network (FBN) is widely used to explore the segregational and integrational organizations of the brain\cite{Park2013Structural}. Numerous approaches have been proposed to identify FBNs in fMRI data, among which data-driven latent variable learning methods based on deep neural networks (DNNs) have proven superb performance compared to those built on conventional shallow matrix factorization models\cite{hjelm2014Restricted,Hu2018Latent,zhang2019Discovering,huang2018Modeling}. The latent variables to learn are either spatial maps that cover brain voxels exhibiting similar temporal fluctuations or time courses that are representative fluctuation patterns, depending on volumetric or time serial input strategy.     

\section{Methods}
\subsection{Formulation of the Framework}
The basic idea to link the ANs in transformed-based NLP models and the BNs in the human brain is to synchronize their temporal activations that respond to the same set of external stimuli (Fig. 1). Let F:S→$Y_a$ represents ANs, and $f_i(S)$ represents the temporal activation of neuron $f_i$ with respect to external stimulus S. Similarly, let G:S→$Y_b$ represents BNs, and $g_j(S)$ denotes the temporal activation of neuron $g_j$ to S. The BN that is anchored by an AN $f_i$ is identified according to Eq. 1. 
\begin{equation}
	\label{eq1}
	{\rm Sync}(f_i,G) = \mathop{\arg\max}\limits_{g_j\in G}\:\delta(f_i,g_j)
\end{equation}
where  $\delta(\cdot)$ is a function measuring the synchronization between the two temporal activations. The AN that is anchored by a BN $g_i$ is identified similarly according to Eq. 2.
\begin{equation}
	\label{eq2}
	{\rm Sync}(g_i,F) = \mathop{\arg\max}\limits_{f_j\in G}\:\delta(g_i,f_j)
\end{equation}
We use the Pearson correlation coefficient (PCC) as a simple but effective $\delta(\cdot)$ to measure the synchronization. We define ANs and BNs, as well as their temporal activations that response to input sequence in the following sections.

\begin{figure}[t]
	\centering
	\includegraphics[width=1.0\linewidth]{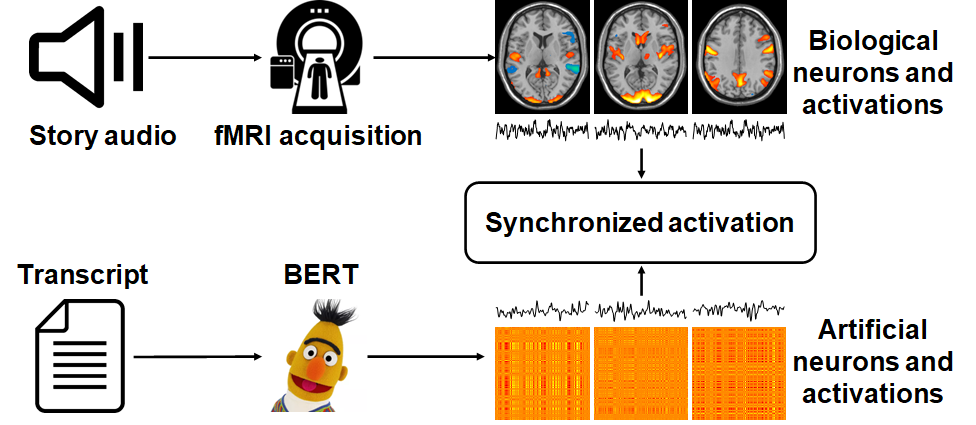}
	% \end{center}
\caption{The framework for coupling artificial neurons in BERT and biological neurons in the brain.}
\label{Figure 1}
\end{figure}

\subsection{ANs and Their Temporal Activations}
A key component in transformer-based NLP model is the multi-head self-attention (MSA) module. The attention score matrix $\mathbf{A}=softmax(\mathbf{Q^TK}/\sqrt{d})$ characterizes how the model attends to different parts of the input (Fig. 2a), where $\mathbf{Q}=\{q_1,q_2,\cdots,q_n\}$ is the query set, $\mathbf{K}=\{k_1,k_2,\cdots,k_n\}$ is the key set, $d$ is the embedding dimension in MSA, and $\textit{n}$ is the number of tokens in the input sequence. After removing the softmax operation for simplification, a single entry in the attention matrix is formulated as $a_{ij}=q_i\cdot k_j=\sum _{1}^{d}q_i \cdot\times k_j$ (Fig. 2b), where $\cdot\times$ denotes element-wise product. It is straightforward that each dimension of the element-wise product of $q_i$ and $k_j$ contributes differently to the dot product and hence attention. Thus, we define each dimension of the query/key vector in the MSA as an individual AN in the BERT model (Fig. 2c). In this way, we can define $\textit{N}_L\times\textit{N}_H\times\textit{d}$ (e.g., 9216 in BERT) ANs in transformer-based NLP models, where $\textit{N}_H$ and $\textit{N}_H$ are the numbers of layers and heads, respectively. 

We then quantify the temporal activation of an AN to external stimuli. Aligning the temporal activations of ANs and BNs is a prerequisite to measure the synchronization between them. To this end, the input text sequence is tokenized and partitioned into subsets according to the temporal resolution (repetition time, TR=1.5s in this study) of fMRI sequence. Let $\{t_1, t_2,\cdots,t_m\}$ denote the $m$ tokens in the $j$-th subset (corresponding to the $\textit{j}$-th time point in fMRI), $\mathbf{Q}_j^{l,h}=\{q_1^{l,h},q_2^{l,h},\cdots,q_m^{l,h}\} $ and $\mathbf{K}_j^{l,h}=\{k_1^{l,h},k_2^{l,h},\cdots,k_m^{l,h}\} $ denote the queries and keys in the $h$-th head and $l$-th layer of BERT, respectively. The $i$-th dimension of the corresponding element-wise product $\mathbf{EP}_j^{l,h,i} \in \mathbf{R}^{m \times m}$ (Fig. 2) measures how a single dimension in the query/key vector (a single AN) contributes to the calculation of attentional relationship among all the $m$ queries and $m$ keys. In other words, the attention matrix can be factorized into $\textit{d}$ independent components (Fig. 2c). Thus, we define the activation of a single AN at time point $\textit{j}$ as the mean of the entries in $\mathbf{EP}_j^{l,h,i}$. The temporal activations of an AN is derived by iterating through all the token subsets (time points), followed by convolution with a canonical hemodynamic response function (HRF) implemented in SPM\footnote{https://www.fil.ion.ucl.ac.uk/spm/} to count for compensation for hemodynamic latency.

\begin{figure}[t]
\centering
\includegraphics[width=0.8\linewidth]{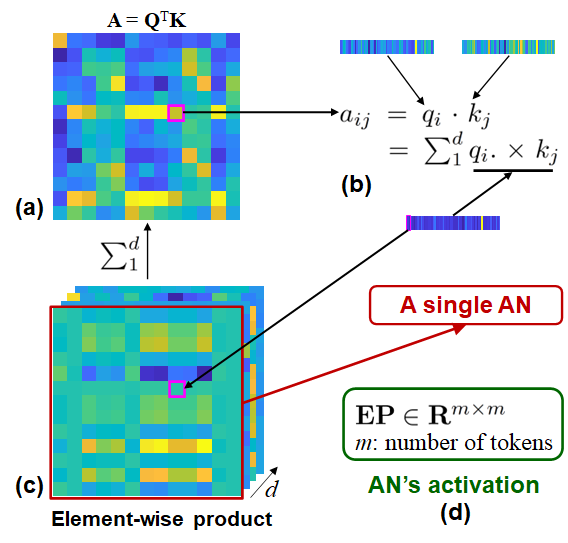}
% \end{center}
\caption{The definitions of AN and its activation in BERT. (a) The simplified attention matrix by removing the softmax operations. (b) The contribution of each dimension in query/key to the attention matrix. (c) The element-wise product of queries and keys. (d) The AN's activation is measured by each dimension of the element-wise product.}
\label{Figure 2}
\end{figure}

\subsection{BNs and Their Temporal Activations}
We treat an FBN as a single BN. We adopt a well-established fMRI analytical approach, namely, the volumetric sparse deep belief network (VS-DBN)\footnote{ https://github.com/QinglinDong/vsDBN} to identify FBNs in fMRI data \cite{dong2019modeling}. In brief, the VS-DBN takes a volume of the fMRI sequence as a feature and each time frame as a sample (Fig. 3a). The VS-DBN model consists of three layers of restricted Boltzmann Machines (RBMs).The first RBM is with $\textit{N}$ visible units, where $\textit{N}$ is the number of valid voxels in the fMRI volume (Fig. 3b). The VS-DBN is trained to discover a set of latent spatial maps, each of which consists of voxels exhibiting similar fluctuation patterns over time and represents the spatial distribution of an FBN. The weights in RBMs are trained layer-wisely with an L1 penalty to enforce sparsity. The linear combination approach that performs successive multiplication of weights from the third to the first RBM (Fig. 3c) is used to generate the global latent variables $\mathbf{W}$. Each column in $\mathbf{W}$ represents a spatial map (Fig. 3d). The number of FBNs is determined by the number of hidden units ($\textit{m}$) in the third RBM layer. The responses of a single hidden unit in the third RBM (Fig. 3e) to the entire input fMRI sequence is the corresponding time series of an FBN and thus is regarded as the temporal activation of an FBN (Fig. 3f).

\begin{figure}[t]
\centering
\includegraphics[width=1.0\linewidth]{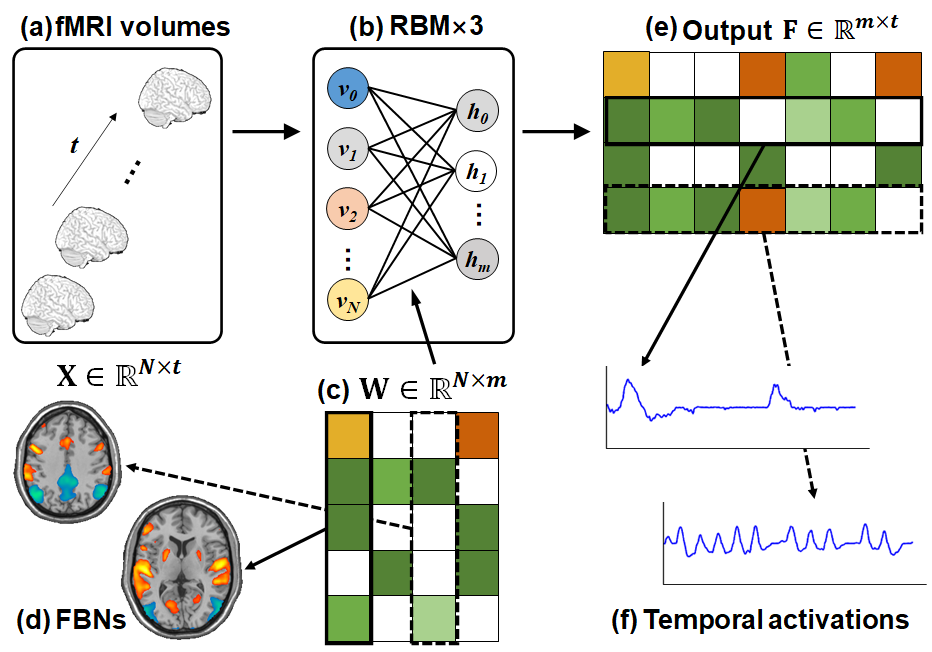}
% \end{center}
\caption{The definitions of BN and its activation in fMRI. (a) The model input is fMRI volumes. (b) The VS-DBN consisits of three RBM layers. (c) The global latent variable $W$. (d) The spatial distribution of FBNs. (e) The temporal activations of FBNs.}
\label{Figure 3}
\end{figure}

\section{Experiments}
\subsection{Dataset and Preprocessing}
We use the “Narratives” fMRI dataset \cite{nastase2021narratives} in this study. The fMRI data were acquired while human subjects listened to 27 diverse naturalistic spoken stories. The "Narrative" fMRI dataset was released with various preprocessed versions. We use the AFNI-smooth version of two fMRI sessions, the “Pie man” (duration 7:20, word count 957, 282 fMRI volumes, spatial resolution  $\rm3\times3\times4 mm^3$, randomly selected 75 subjects from the 82 subjects in total) and “Shapes” (duration 6:45, word count 910, 270 fMRI volumes, spatial resolution $\rm3\times3\times4 mm^3$, all the 59 subjects). For the fMRI sequence of a subject, the volumes before the onset and after the end of the story stimuli are discarded. The time series of each voxel is normalized to have zero mean and unit standard deviation.

The "Pie man" is a story about a journalist writing reports of a man with supernatural abilities. The "Shapes" is about using two-dimensional geometric shapes to tell a story of a boy who dreams about a monster. It is noteworthy that the spoken story "Shapes" is intended to convey intentionality in the animated shapes \cite{nastase2021narratives}. The spoken story stimuli are released with time-stamped word-level transcripts. It helps to align text tokens with fMRI volumes. We tag the part-of-speech (15 categories of part-of-speech, and two additional categories of [CLS] and [SEP]) for each token in the transcripts via spaCy\footnote{https://spacy.io}.

% \subsection{Implementation Details}
\subsection{Implementation Details}
We use the pre-trained BERT model (BERT-base, 12 layers, 12 heads, hidden size 768) maintained by HuggingFace\footnote{https://huggingface.co/docs/transformers/model-doc/bert} in this study to implement the proposed framework. Each of the tokenized transcripts is separated into three disjoint segments by balancing the limit of maximum number of tokens (512) in BERT and the completeness of sentences. There are 505-468-254/411-416-268 tokens in the three segments of “Pie man”/“Shapes”. 

All the volumes in the two sessions are aggregated to train the VS-DBN with the following parameters: 512/256/128 hidden units in the 1\textsuperscript{st}/2\textsuperscript{nd}/3\textsuperscript{rd} RBM layer, Gaussian (zero-mean and a standard deviation of 0.01) initialization, learning rate 0.001/0.0005/0.0005, batch-size 20, L1 weight-decay rate 0.001/0.00005/0.00005, 100 training epochs, batch normalization. Model training is performed on a work-station with 10 GeForce 1080Ti GPUs. The FBNs in the two fMRI sessions share the same set of spatial maps but have their own temporal activations for each subject. The session-specific temporal activations are averaged over subjects. 

\section{Results}
\subsection{Synchronized Activations between ANs and BNs}
We identify the BN that is anchored by an AN using Eq. 1. The distribution of the ANs’ maximum PCC to BNs are shown in Fig 4(a). The PCCs are statistically significant ($p < 0.01$, permutation test with 5000 randomizations, FDR corrected) for 9176/9097 (99.57\%/98.71\%) ANs in “Pieman”/“Shapes”. The average PCC of ANs in each layer (Fig. 2b) approximately increases linearly, indicating the ANs in deeper BERT layers better synchronize to BNs. Meanwhile, the number of ANs with PCC below 0.25 on each layer exhibits a decreasing trend in both sessions (Fig. 2c-d).  

\begin{figure}[t]
\centering
\includegraphics[width=1.0\linewidth]{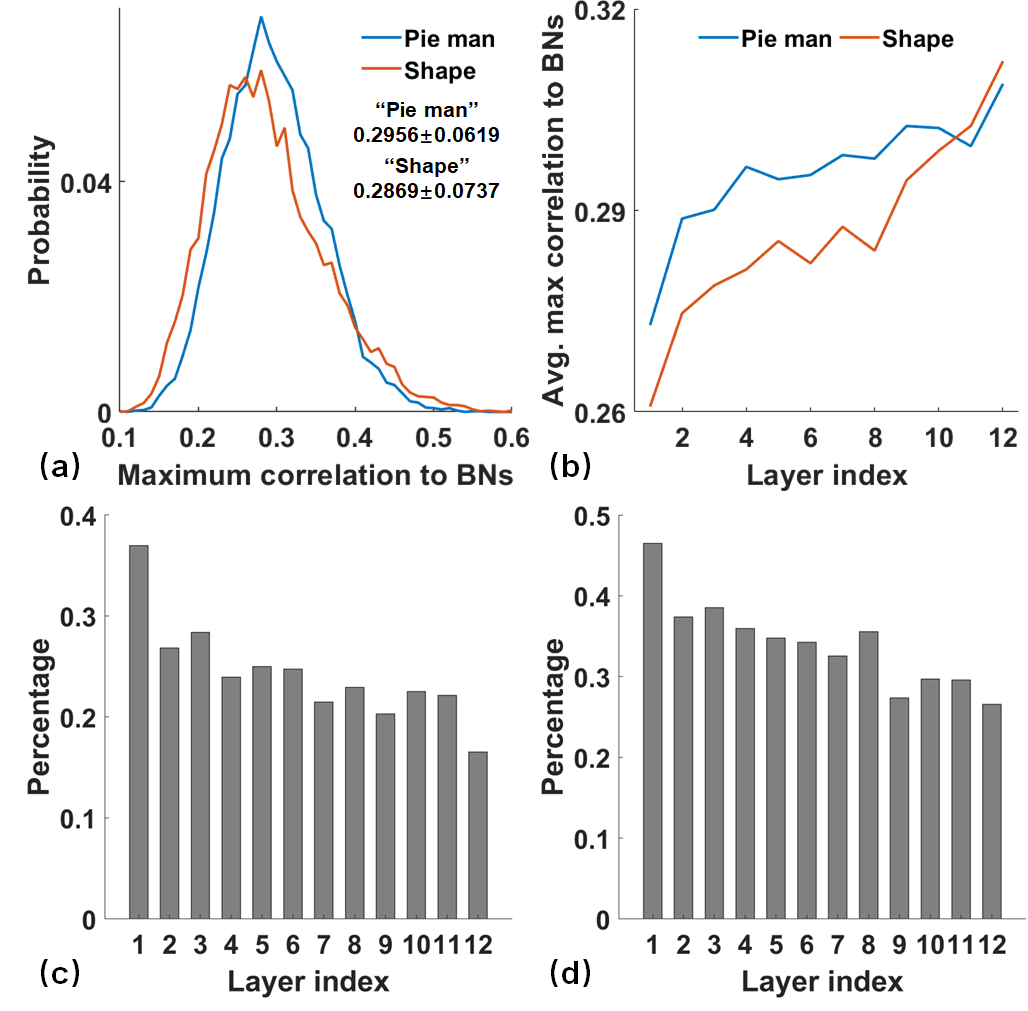}
% \end{center}
\caption{(a) The distribution of the ANs’ maximum PCC to BNs. (b) The average PCC in each layer. (c-d) The percentage of ANs with PCC below 0.25 in each layer.}
\label{Figure 4}
\end{figure}

We identify the AN that is anchored by a BN using Eq. 2. The PCCs (mean±std: 0.4364±0.0473 in “Pie man”, Fig. 5a; 0.4422±0.0592 in “Shapes”, Fig. 5b) are statistically significant ($p=0$, permutation test with 5000 randomizations) for all the 128 BNs in both sessions. The number of ANs that are anchored by BNs in each layer (Fig. 5c-d) show that ANs on deeper BERT layers (10-12) are more frequently anchored by BNs compared to the ones in shallower layers.

\begin{figure}[t]
\centering
\includegraphics[width=1.0\linewidth]{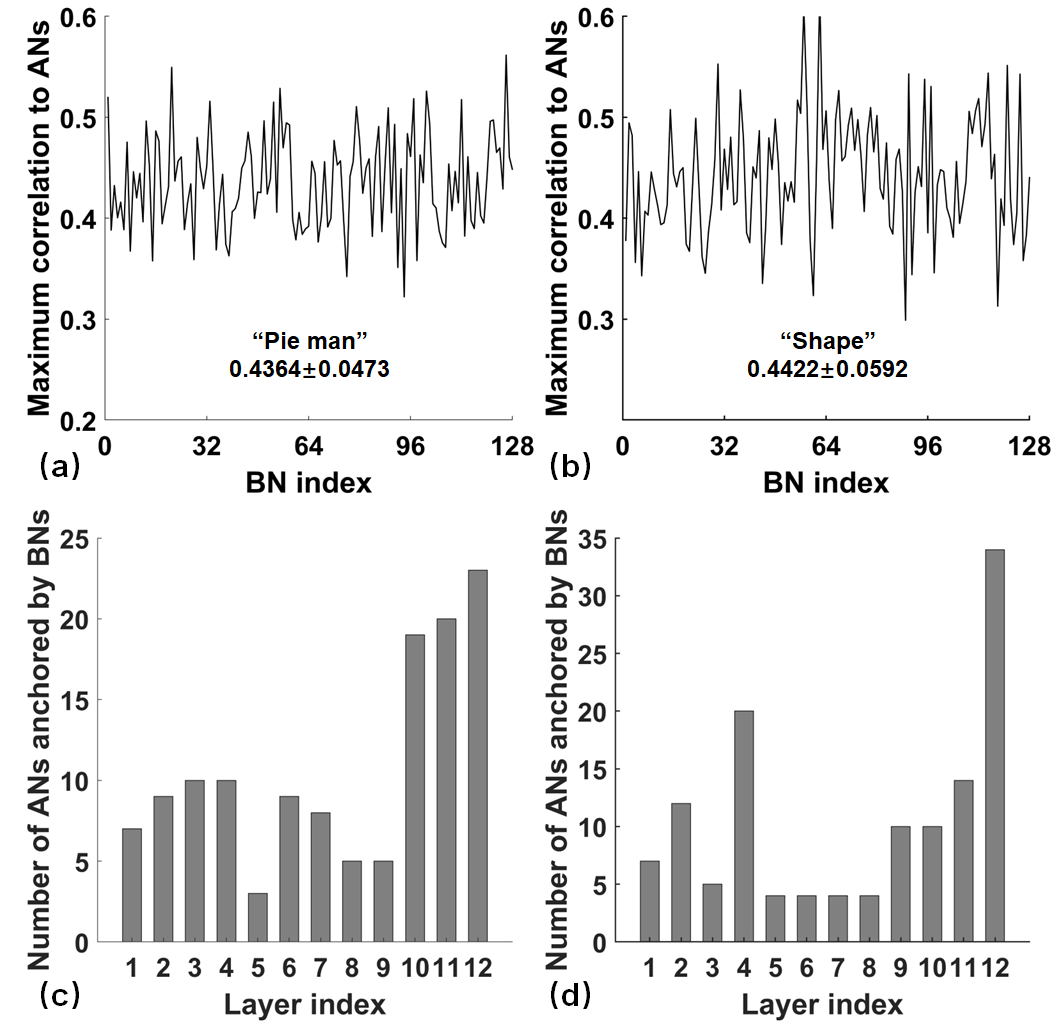}
% \end{center}
\caption{(a-b) The BNs’ maximum PCCs to ANs in the two sessions. (c-d) The number of ANs that are anchored by BNs in each layer.}
\label{Figure 5}
\end{figure}

\subsection{The Most Frequently Anchored BN by ANs}

With a PCC threshold 0.25, we count the times that a BN is anchored by ANs and identify the BN that is anchored the most frequently in the two sessions separately. The BN \#89/\#57 is with the largest number of anchored ANs in “Pie man”/“Shapes” (Fig. 6a). The BN \#89 (Fig. 6b) mainly encompasses activations of the left-lateralized Broca’s area and its counterpart on the right hemisphere (yellow arrows), bilateral visual word fusiform areas (VWFA, red arrow) that are consistently active in reading, and bilateral middle occipital gyri (green arrow). The Broca's area is well-known as one of the core regions in the language network in the human brain. The VWFA has a specific role in decoding written forms of words \cite{dehaene2011The} while recent studies have shown a multiplex model of VWFA function characterized by distinct circuits for integrating language and attention \cite{chen2019vwfa,Sani2021vwfa}. In addition, the activation of the primary visual cortex in language processing has been widely observed and discussed \cite{Seydell2021visual,Pennartz2019Towards}. The BN \#57 (Fig. 6c) encompasses the working memory network (WM, red arrows), the salience network (green arrows) and the default mode network (DMN, yellow arrows). 

\begin{figure}[t]
\centering
\includegraphics[width=1.0\linewidth]{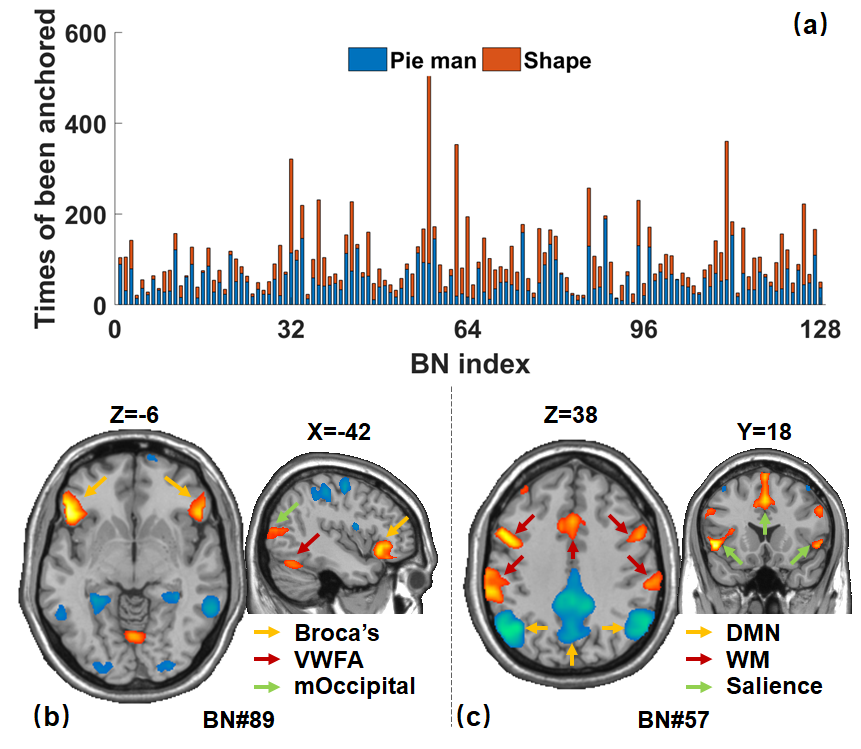}
% \end{center}
\caption{(a) The number of ANs that are anchored by a BN. (b) The spatial distribution of BN\#89. (c) The spatial distribution of BN\#57.}
\label{Figure 6}
\end{figure}

We then interpret this difference in a linguistics context. Using the part-of-speech tagging of tokens, we tag the part-of-speech for an AN. More specifically, for each AN we identify 500 query-key pairs that have top element-wise product in each segment of the tokenized transcript, resulting in 1500 query-key pairs. The 17 part-of-speech tags of tokens determine $17\times17$ part-of-speech categories for query-key pairs. We then count the number of query-key pairs falling into each category. An AN is tagged by the category having the most query-key pairs. 

We summarize the distribution of AN tags in Fig. 7(a-b) for the two sessions, and their difference (“Pieman”–“Shapes”) in Fig. 7(c). More ANs in “Pie man” are tagged as “Determiner$\leftrightarrow$Noun”, “Verb$\leftrightarrow$Noun” and “Verb$\leftrightarrow$Verb” (red). In comparison, more ANs in “Shapes” are tagged as “Pronoun$\leftrightarrow$Pronoun”, “Punctuation$\leftrightarrow$Punctuation”, “Pronoun$\leftrightarrow$Punctuation”, “Pronoun$\leftrightarrow$Verb” (blue). Thus, it is reasonable that the most frequently anchored BN in “Pie man” is \#89, which falls into the functionally specialized brain regions that focus on lexicons/concepts; and the most frequently anchored BN in “Shapes” is \#57, which falls into the domain-general brain regions that are sensitive to syntactic/context evolving over time. For example, fMRI studies using dynamic naturalistic stimuli (including the auditory story in this study) suggested that the DMN plays a central role in integrating incoming extrinsic information (temporarily stored in the WM) with prior intrinsic information over relatively long timescales to form context dependent models \cite{yeshurun2021default}.

\begin{figure}[t]
\centering
\includegraphics[width=1.0\linewidth]{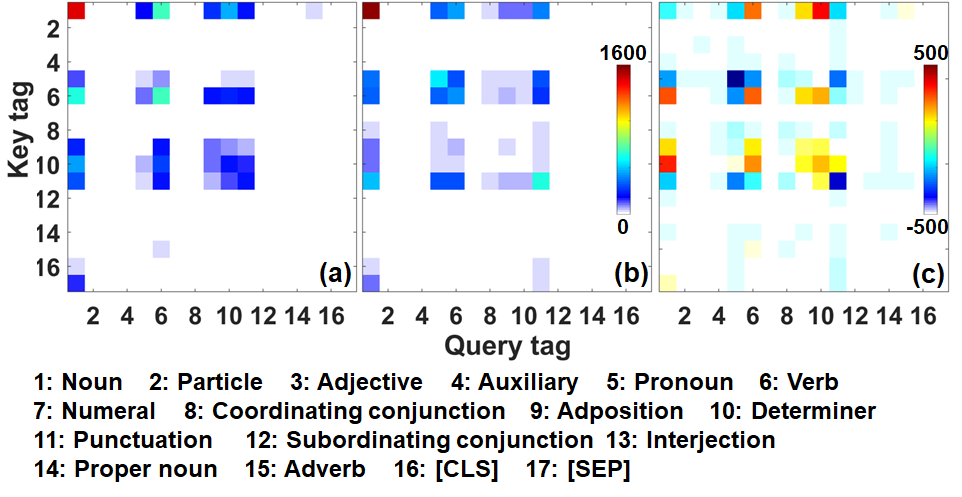}
% \end{center}
\caption{The distributions of part-of-speech tags of ANs in “Pie man” (a) and “Shapes” (b). (c) The difference between (a) and (b).}
\label{Figure 7}
\end{figure}

\subsection{The Best Synchronized BN-AN Pair in Each Layer}
We identify the BN-AN pair with the largest PCC in each layer, as shown in Fig. 8 for the “Pieman” session. The corresponding BNs in the first four layers commonly cover activations in the auditory-language network (including the Heschl’s gurus, the Broca’s area and the Wernick’s area) and its counterpart on the right hemisphere, and the dorsal/ventral visual areas. The BNs in the middle layers of L5-L7 show common activations in the primary dorsal visual areas and the somatosensory network. The BN in layer 8, where the PCC is maximized globally, exhibits coactivations of the language network and DMN. In deeper layers of L9-L12, common activations in the high-level ventral visual cortex (mainly the fusiform gyrus) and some subcortical/limbic structures (e.g., putamen, caudate and insula, responsive to high-order cognitive processes such as emotion) are observed. 

\begin{figure}[t]
\centering
\includegraphics[width=1.0\linewidth]{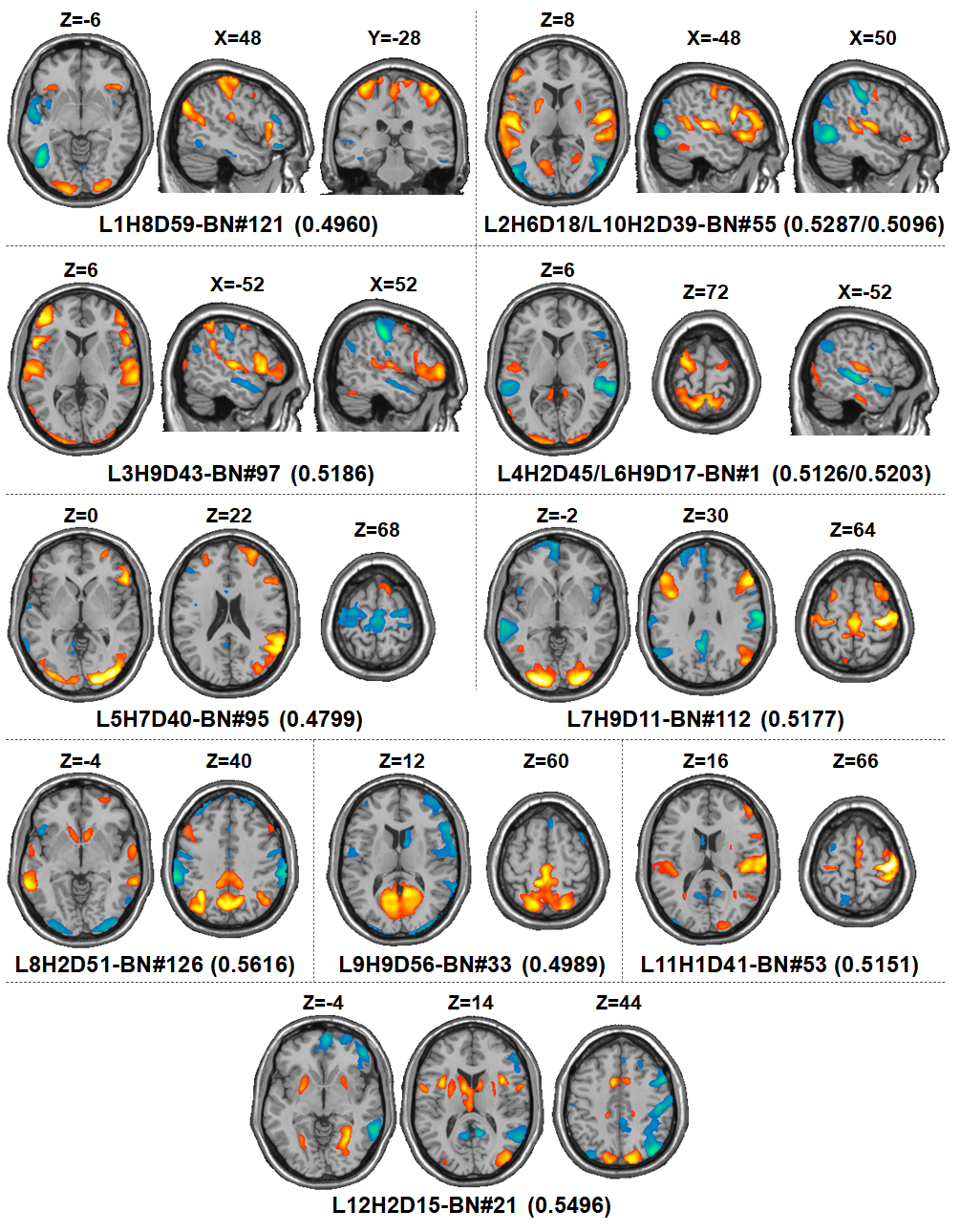}
% \end{center}
\caption{The best synchronized BN-AN pair in each layer in “Pie man"). The number in the brackets is the PCC. L1H8D59 represents dimension 59 in head 8 in layer 1.}
\label{Figure 8}
\end{figure}

In “Shapes” (Fig. 9), the BNs in multiple layers including L1, L2, L5, L7, L8 and L10 are identical (\#57) and are associated with domain-general brain regions including the DMN, WM and salience network as illustrated in Fig. 6(c). We also observe the activation/deactivation of the DMN in the BNs in L3, L6, L11 and L12. The activations of the language network and its counterpart on the right hemisphere are present in the BNs in L4 and L9.

\begin{figure}[t]
\centering
\includegraphics[width=1.0\linewidth]{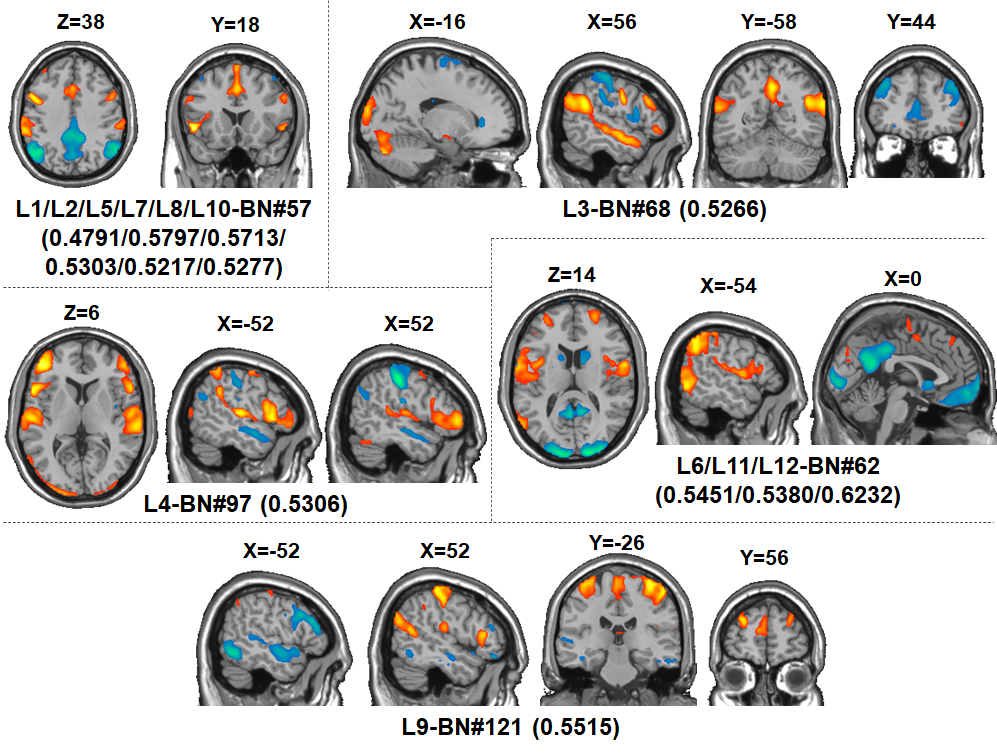}
% \end{center}
\caption{The best synchronized BN-AN pair in each layer in “Shapes"). The number in the brackets is the PCC.}
\label{Figure 9}
\end{figure}

We further look into the details of the internal representations of the corresponding ANs in each layer in terms of part-of-language tags. We illustrate several exemplar BN-AN pairs in “Pie man”. The part-of-speech tag distribution for the rest of BN-AN pairs in "Pie man" is referred to Supplementary Fig. 1, and those in "Shapes" are referred to Supplementary Fig. 2. The first example is the globally best synchronized BN-AN pair (AN\#L8H2D51-BN\#126). The part-of-speech tag distribution of those query-token pairs is shown in Fig. 10(a). The “Noun→Pronoun”, “Noun→Punctuation”, “Noun→Coordinating conjunction”, “Verb→Punctuation”, “Verb→Pronoun” are among the top six tag categories. Queries of Noun (30.20\%) and Verb (20.27\%) and keys of Pronoun (23.07\%) and Punctuation (21.53\%) are predominant. Thus, the resolution of pronoun may recruit domain-general brain regions (i.e., the DMN here) that assemble memory retrieval \cite{li2020modeling}. 

The second example is the BN-AN pair (AN\#L1H8D59-BN\#121) in the first layer. The distribution (Fig. 10b) shows that the category of Noun-Noun dominates (13.53\%) the tags of query-key pairs. And intriguingly, the activation of functionally specialized auditory-language network is observed. The third example is the BN-AN pair (AN\#L3H9D43-BN\#121) on the third layer. The tags distribute (Fig. 10c) more diversely compared to those in the two examples presented above, indicating that the AN is associated with general language processing and activates the primary auditory-language network. 

\begin{figure}[t]
\centering
\includegraphics[width=1.0\linewidth]{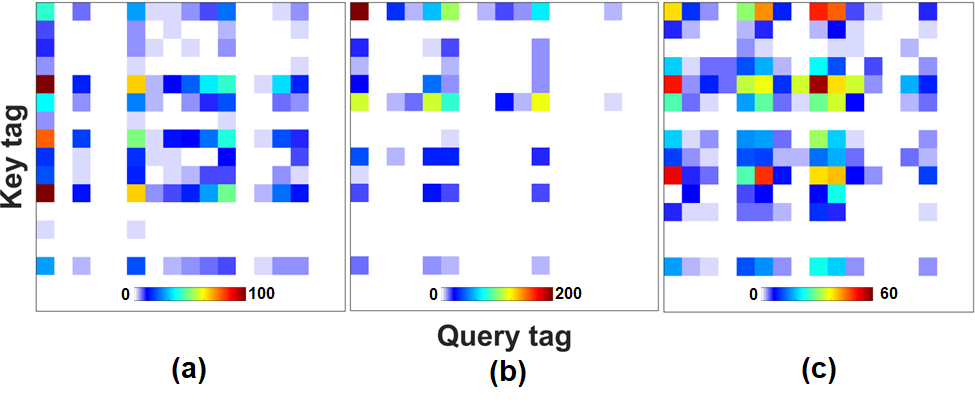}
% \end{center}
\caption{The distribution of part-of-speech for three exemplar ANs in “Pie man”. }
\label{Figure 10}
\end{figure}

For an ease reference, we also use the text attention heatmap visualization tool\footnote{https://github.com/jiesutd/Text-Attention-Heatmap-Visualization} to provide visualizations of the top 1500 query-key pairs overlayed on the transcript for the AN-BN pairs in each layer for the two sessions in Supplementary Fig. 3 and Fig. 4, respectively. In the visualizations, red color encodes queries and blue color encodes keys. The color is coded according to the magnitude of the element-wise product of queries and keys.

\section{Conclusion}
In the current study, we proposed a framework to couple artificial neurons (ANs) in transformer-based NLP models and biological neurons (BNs) in the human brain. Compared to existing studies that treat each layer in the model as an AN, we improved the granularity of ANs by defining each dimension in the embedding in each layer as a single AN. Meanwhile, we defined functional brain networks (FBNs) as BNs to reveal complex functional interactions in the brain when it was exposed to naturalistic linguistic stimuli. Then, the correspondences between ANs and BNs were established by maximizing their temporal activations. Our experimental results demonstrated that the temporal activations of the ANs and BNs defined in this study were significantly synchronized. We also partly demonstrated that the ANs carry meaningful linguistic features and the BNs that anchored by ANs were interpretable in a neurolinguistic context. The framework proposed in this study may serve as a brain-based test-bed to evaluate and interpret transformer-based NLP models.  

The present study is considered with respect to some limitations. First, we used the pre-trained bidirectional BERT model. However, the human brain attends unidirectionally. Meanwhile, the parameters of the pre-trained BERT would be updated by downstream tasks. Thus, it would be interesting in future studies to explore whether there are consistent AN-BN coupling patterns across different NLP models, for example unidirectional GPT3 \cite{GPT3} and the ones that are fine-tuned by downstream tasks. Second, a strong baseline control experiment (e.g., fMRI acquired using shuffled spoken stories) is further necessary to confirm the coupling between ANs and BNs. Third, we used the mean entries in the element-wise product of query and key to measure the activation of ANs, which may face the risk of information loss. Some alternatives such as maximum/minimum are worth to try. At last, the neurolinguistic interpretation is bounded to a limited number of ANs and BNs and by a single linguistic attribute of part-of-speech tagging. In the future, it is desirable to perform systematical analysis to explore the linguistic and semantic information carried by the ANs and link them to their symbiosis of BNs.

\section{Acknowledgments}
This work was partly supported by National Key R\&D Program of China (2020AAA0105701), National Natural Science Foundation of China (62076205, 61936007 and 61836006). 

\bibliography{aaai23}

\end{document}